\documentclass[letterpaper, 10 pt, conference]{ieeeconf}  

\IEEEoverridecommandlockouts                              

\usepackage{graphicx}
\usepackage{hyperref}                       
\graphicspath{{figures/}}
\usepackage{tikz}							
\usepackage{pgfplots}						
\pgfplotsset{compat=newest}					
\usetikzlibrary{positioning}				
\usepackage{siunitx}						
\usepackage{booktabs}						
\usepackage{mathtools} 						
\usepackage{nicefrac} 						
\usepackage{multirow} 						
\usepackage{subcaption}						
\usepackage{cite}							
\usepackage[ruled]{algorithm2e}				
\usepackage{amssymb}						

\newcommand{\ego}{ego-vehicle }				
\newcommand{\egowo}{ego-vehicle}			
\newcommand{\tabelle}{Table~}				
\newcommand{\abbildung}{Fig.~}				
\newcommand{\gleichung}{Equation~}			
\newcommand{\algo}{Algorithm~}   			
\renewcommand{\vec}[1]{\mathbf{#1}}			

\definecolor{TUMGreen}{RGB}{162, 173, 0} 
\definecolor{TUMOrange}{RGB}{227, 114, 34}
\definecolor{TUMGray}{RGB}{156, 157, 159}       
\definecolor{TUMGray3}{RGB}{217, 218, 219}       
\definecolor{TUMDiag1}{RGB}{105, 8, 90}  
\definecolor{TUMDiag14}{RGB}{249, 186, 0} 
\definecolor{TUMDiag9}{RGB}{0, 124, 48}       
\colorlet{green}{TUMGreen}  
\colorlet{red}{TUMOrange}

\SetCommentSty{mycommfont} 					

\makeatletter 
\g@addto@macro{\@algocf@init}{\SetKwInOut{Parameter}{Parameter}} 
\makeatother

\newlength\figureheight             		
\newlength\figurewidth 						

\graphicspath{{figures/}}
\makeatletter
\def\input@path{{figures/}} 
\makeatother

\usetikzlibrary{external}			
\tikzset{external/system call={pdflatex -shell-escape -buf-size=10000000 -extra-mem-top=100000000 -halt-on-error -interaction=batchmode -jobname "\image" "\texsource"}} 

\usetikzlibrary{external}
\usepackage{textcomp}
\usepackage{lipsum}
\newcommand\copyrighttext{%
	\footnotesize \textcopyright 2020 IEEE.  Personal use of this material is permitted.  Permission from IEEE must be obtained for all other uses, in any current or future media, including reprinting/republishing this material for advertising or promotional purposes, creating new collective works, for resale or redistribution to servers or lists, or reuse of any copyrighted component of this work in other works.
}
\newcommand\copyrightnotice{%
	\tikzset{external/export=false}
	\begin{tikzpicture}[remember picture,overlay]
	\node[anchor=south,yshift=10pt, xshift=10pt] at (current page.south) {\fbox{\parbox
			{\dimexpr\textwidth-\fboxsep-\fboxrule\relax}{\copyrighttext}}};
	\end{tikzpicture}%
	\tikzset{external/export=true}
}

\title{\LARGE \bf
	Automatic Generation of Road Geometries to Create Challenging Scenarios for Automated Vehicles Based on the Sensor Setup*
}

\author{Thomas Ponn$^{1}$, Thomas Lanz$^{1}$ and Frank Diermeyer$^{1}$
\thanks{*The research project was funded and supported by T\"UV S\"UD Auto Service GmbH.}
\thanks{$^{1}$Thomas Ponn, Thomas Lanz and Frank Diermeyer are with the Institute of Automotive Technology, Technical University of Munich, 85748 Garching, Germany
        {\tt\small \{ponn,diermeyer\}@ftm.mw.tum.de},{\tt\small ga87veq@mytum.de}}%
}

\begin{document}

\maketitle \copyrightnotice
\thispagestyle{empty}
\pagestyle{empty}
%
%
\begin{abstract}

For the offline safety assessment of automated vehicles, the most challenging and critical scenarios must be identified efficiently. Therefore, we present a new approach to define challenging scenarios based on a sensor setup model of the \egowo. First, a static optimal approaching path of a road user to the \ego is calculated using an A* algorithm. We consider a poor perception of the road user by the automated vehicle as optimal, because we want to define scenarios that are as critical as possible. The path is then transferred to a dynamic scenario, where the trajectory of the road user and the road layout are determined. The result is an optimal road geometry, so that the \ego can perceive an approaching object as poorly as possible. The focus of our work is on the highway as the Operational Design Domain (ODD). 

\end{abstract}


\section{INTRODUCTION}
\label{sec:introduction}

For the introduction of automated vehicles (AV) where the vehicle is responsible for performing the driving task (automation level 3 and higher according to SAE \cite{SAEJ3016.2018}), their safety must be assessed. The aim is to exceed a minimum level of safety that has not yet been defined. However, the procedure for the efficient and economical implementation of this safety verification is an open problem in automotive engineering. A distance-based procedure based exclusively on real tests is no longer economically feasible due to the enormous effort involved. According to \textsc{Wachenfeld and Winner} \cite{Wachenfeld.2016}, approximately 6.6 billion kilometers are required under representative conditions to determine the safety level of an Autobahn chauffeur on German motorways.

In real-life road traffic, the proportion of comparatively simple situations that do not provide any added value for proof of safety is high, which is why the so-called scenario-based approach (like in the German funded project PEGASUS \cite{DeutschesZentrumfurLuftundRaumfahrte.V..2018}) is restricted specifically to important and relevant scenarios. While the number of tests performed by virtual simulation is rising and the quality and performance of simulation is improving, even with this tool only a limited number of tests can be performed. But due to continuous parameters and an open parameter space, theoretically an infinite number of test scenarios can be defined \cite{Althoff.2014, Huang.2017, Junietz.2018, Menzel.2018}.

From this infinite number of possible test cases, one must select a manageable amount that is especially meaningful. These particularly good and challenging test cases are also called edge or corner cases. This is explicitly important for the type approval of automated vehicles because only a very limited number of tests can be carried out for the economic implementation of the type approval. An automated vehicle consists of several modules \cite{Amersbach.2019} and test cases can be defined that represent a specific challenge for a certain module. In the context of this publication, the most challenging scenarios for the perception module of AV, especially for the use case of driving on German motorways, shall be defined. Because there are various possibilities in the implementation of an AV (e.g. different sensor setups), the efficient selection of challenging scenarios must be system-specific. 

This contribution therefore presents a novel approach for a system-specific definition of particularly challenging scenarios for the perception module of AV. This work is based on a previously published work of the author\cite{Ponn.2019b}. Additionally, the presented approach is part of the overall methodology of the author to identify relevant scenarios for the type approval of AV presented in \cite{Ponn.2019}. 


\section{RELATED WORK}
\label{sec:related_work}
This chapter provides an overview of different techniques for defining and identifying scenarios for testing AV. Subsequently, it is shown in which aspects the present work represents an extension to the current state of the art. 

\subsection{Testing of Automated Vehicles}
\label{subsec:Subsection_1}
For the assessment of AV, the scenario-based approach already mentioned in Section \ref{sec:introduction} is used. The basic idea of the scenario-based approach is to verify the most relevant and critical scenarios that could occur instead of all possible ones so that the overall driving distance can be significantly reduced \cite[p.46]{Mazzega.2016}. The basic assumption is that the greater part of everyday driving is irrelevant, e.\,g. driving on a straight and empty highway.

There are various approaches in literature on how to generate relevant scenarios:
\begin{LaTeXdescription}
	\item[Traffic analysis:] For example, accident databases can be used to identify especially critical scenarios when analyzing real traffic situations \cite{Fahrenkrog.2019, So.2019,Stark.2019}. In addition, a criticality metric can be used to filter out critical situations from test drives that represent relevant scenarios in which no accident occurred \cite{Junietz18}. The exposure of individual scenarios can also be taken into account in order to make a more precise statement about the safety level of the vehicle \cite{Gelder.2017}. 
	\item[Combinatorial testing and Design of Experiments:] For simple systems, an n-wise combination of the parameters (i.e. each parameter value is combined with each parameter value) is still feasible, but for higher degrees of automation this is no longer feasible even using virtual simulation \cite{Ponn.2019c}. A reduced number of combinatorial test cases based on scenario importance is part of \cite{Huang.2018}. \textsc{Schuldt et al.} \cite{Schuldt.2018} discretize value-continuous parameters into equivalence classes and therefore reduce the number of test cases for a necessary test coverage. 
	\item[Search based methods:] All search-based approaches presented in the following have in common that they require a (simplified) model of the driving function. \textsc{Ben Abdessalem et al.} \cite{BenAbdessalem.2018} use Evolutionary Algorithms to generate critical scenarios for a vision-based automatic emergency braking system. In \cite{Koren.2018} Monte Carlo Tree Search and Deep Reinforcement Learning are used to find scenarios that lead most likely to a system failure. In order to define critical scenarios, Evolutionary Algorithms are used in \cite{M.Klischat.2019}. Thereby the safe drivable area, i.\,e. the solution space for the planning module of the automated vehicle, is minimized. Differential Evolution and Particle Swarm Optimization are used in \cite{H.Beglerovic.2017} to generate critical scenarios for a Emergency Braking Assist. For this purpose, surrogate models of the overall system are implemented to reduce the computation time during optimization.
	\item[Challenging scenarios:] Scenarios that are particularly difficult to master are referred to as challenging scenarios. The basic assumption here is that an increased difficulty of the test case results in an increased probability of the occurrence of system failures. \textsc{Gao et al.} \cite{Gao.2019} define a complexity index based on the Analytic Hierarchy Process that characterizes the traffic situation based on the difficulty. In addition, it is shown that a Lane Departure Warning System fails more frequently in more complex situations. \textsc{Wang et al.} \cite{Wang.2018} propose a new method to determine the complexity of a traffic scene by the quantified road semantic complexity and the traffic element complexity. The former evaluates the static environment and the latter the dynamics of other road users according to their difficulty for an AV. However, the presented metric is not validated.
\end{LaTeXdescription}

For reasons of completeness, the formal methods will be presented as an alternative concept to testing, although this approach will not be discussed in the further course of the work. The verification of the system by formal methods as for example in \cite{Althoff.2014, Kamali.2017, Koopman.2019, Mehmed.2019, N.Arechiga.2019, ShalevShwartz.2017} aims at mathematically proving the safety of AV. If formal methods without assumptions and restrictions are applicable to the overall system, any kind of simulated or real test would be obsolete. However, so far there is no procedure that achieves this at the overall system level and therefore tests will also be necessary in the future to demonstrate the safety of AV.

\subsection{Contributions}
\label{subsec:contributions}
The first question to be answered is why the existing procedures are not sufficient. The methods of traffic analysis require a large amount of data, which is very cost-intensive to generate. The problem with the use of accident databases is that these contain almost exclusively accidents of human drivers. A direct transfer to AVs is not possible. Combinatorial tests have the disadvantage that they do not reduce the number of tests to the required extent. Search-based methods have the disadvantage that a simulation model of the entire system is necessary, with which a large number of simulations must be carried out. On the one hand, high-fidelity simulation models with high computational effort must be used for a realistic evaluation of safety. On the other hand, when AVs are certified by a technical service, not all simulation models may be available for the technical service and search-based approaches may not be applicable. The current literature with regard to challenging scenarios focuses mainly on the evaluation of existing scenarios and not on the definition of new challenging scenarios. Therefore, a method that actively defines challenging scenarios is presented here. The focus will be on the perception of the system and the approach is especially suitable for technical services to support the certification of automated vehicles without having all simulation models of the system available.

\begin{figure*}[]
	\centering
	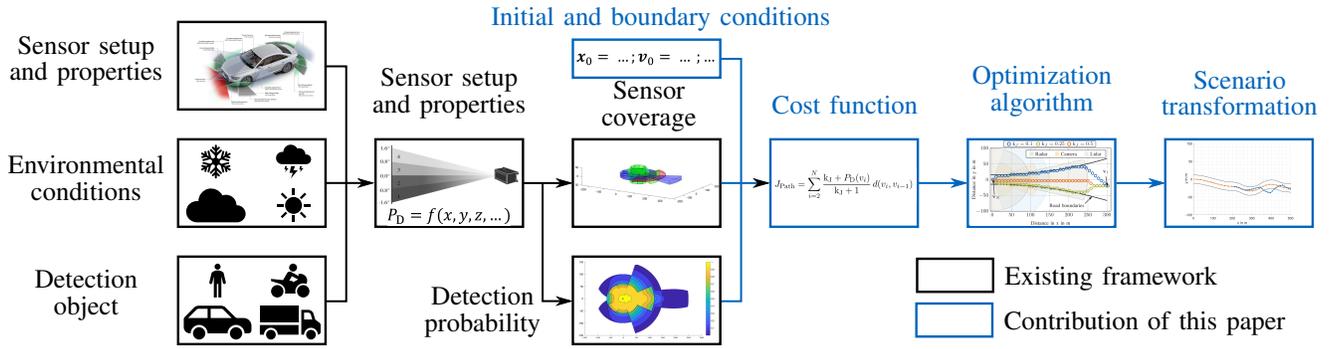
	\caption{Connection between already published framework \cite{Ponn.2019b} and the contribution of the current publication.}
	\label{fig:overallApproach}
	\vspace{-15pt}
\end{figure*}

The basis for the method presented here is an existing framework from our previous work on the calculation of a 3D grid with detection probabilities (\abbildung \ref{fig:overallApproach}). Based on the used sensor setup, environmental conditions and the object to be detected, the sensor coverage and a three-dimensional grid with detection probabilities are calculated with phenomenological sensor models. With the existing framework, weak spots of the sensor setup can already be identified and the focus of the safety assessment of the vehicle can be set on these weak spots. Building on this, the present paper shows how a path finding algorithm can be used to determine a worst case approaching path of an object towards the \ego and to derive an optimal road geometry for this test scenario. The worst case approaching path is the path with which an object must approach an \ego in order to be perceived as poorly as possible. This constellation represents the greatest challenge for the perception of the vehicle and can lead to critical scenarios that are of crucial importance for the safety assessment. 

\section{METHODOLOGY}
\label{sec:methodology}

This chapter describes the individual blocks drawn in blue in \abbildung \ref{fig:overallApproach}. A 3D grid with the detection probabilities $P_\mathrm{D}$ serves as the basis. This means that a probability is assigned to each node in the grid with which an object (e.g. another vehicle) is correctly detected by the \ego under previously defined environmental conditions. All nodes outside the field of view of the sensors have a detection probability of zero. The basic idea is to iterate through the nodes of the grid in order to find the optimal approaching path $\vec{x}_\mathrm{oap}$ where the cost between the start node $v_1$ (start position of the object\footnote{The object approaching the \ego is further referred to as the challenger.}) and the end node $v_N$ (position of \egowo) is minimal. The detection probability $P_\mathrm{D}$ and the length of the path $d$ are considered as assessment quantities of $\vec{x}_\mathrm{oap}$. We want to define particularly challenging scenarios and therefore the path should be as short as possible so that the required reaction time of the \ego to the challenger is as short as possible. At the same time $P_\mathrm{D}$ of $\vec{x}_\mathrm{oap}$ should be minimal so that the probability is increased that the \ego detects the challenger late. This means that a path as short as possible with a low detection probability is optimal for our application.
\subsection{Cost Function}
\label{sub:costFunc}
First, we need a mathematical representation of the distance $d$ and the detection probability $P_\mathrm{D}$. For $d$, the Euclidean distance between the current node $v_j$ and the previous node $v_{j-1}$ is used. For the detection probability $P_\mathrm{D}(v_j)$, the existing values of the grid from our previous work in \cite{Ponn.2019b} are used for each node $v_j$. It is advantageous for the cost function if the ratio between the detection probability $P_\mathrm{D}(v_j)$ and the distance $d(v_j,v_{j-1})$ can be adjusted. This is done with the weighting factor $k_\mathrm{J}$. \gleichung \ref{eq:J} shows the cost function $J_\mathrm{Node}$ that describes the cost of the approaching path between two nodes $v_j$ and $v_{j-1}$. 
\begin{equation}
J_\mathrm{Node} = \dfrac{\mathrm{k}_\mathrm{J} + P_\mathrm{D}(v_j)}{\mathrm{k}_\mathrm{J}+1} \, d(v_j,v_{j-1})\\
\label{eq:J}
\end{equation}
Note that for all $\mathrm{k}_\mathrm{J}$ the distance $d$ is taken into account, since also diagonal steps are possible in the orthogonal grid and these are longer than horizontal and vertical steps. The total cost of the path $J_\mathrm{Path}$ consisting of $N$ nodes can be calculated according to \gleichung \ref{eq:JAll}.
\begin{equation}
J_\mathrm{Path} = \sum \limits_{j=2}^N \dfrac{\mathrm{k}_\mathrm{J} + P_\mathrm{D}(v_j)}{\mathrm{k}_\mathrm{J}+1} \, d(v_j,v_{j-1})\\
\label{eq:JAll}
\end{equation}
The choice of $\mathrm{k}_\mathrm{J}$ has a considerable influence on the result, which is examined in more detail in Section \ref{sec:results}. The higher $\mathrm{k}_\mathrm{J}$ in \gleichung \ref{eq:JAll}, the less $P_\mathrm{D}$ taken into account and thus the greater focus given to the shortest possible approaching path. Preliminary investigations in \cite{Lanz.2019} have revealed that values in the range of $0.1-1$ are promising for $\mathrm{k}_\mathrm{J}$. 
\subsection{Optimization Algorithm}
\label{sub:optiAlgo}
This section describes the calculation of the path through the three-dimensional grid with the lowest cost. It also shows how to reduce the number of valid nodes in the search space. Within our work, two different optimization algorithms are examined. First, the A-star (A*) algorithm, a heuristic search algorithm \cite[chap.\,6.3.2]{Ertel.2017}, and second, the Ant Colony Optimization with problem-specific adjustments, which is one of the nature-inspired path finding algorithms \cite{Du.2016}. In our preliminary investigations \cite{Lanz.2019}, the A* algorithm proved to be superior to the Ant Colony Optimization in terms of run-time, quality and simple parameter selection. In this paper, we will therefore only discuss the A* algorithm. 

The following description of the functionality of the A* algorithm is based on \cite[chap.\,6.3.2]{Ertel.2017}. One characteristic of the A* algorithm that indicates its suitable use is that the optimal path is always found between the start and end nodes, if one exists. The cost of the path for the A* algorithm via $v_j$ is determined by a heuristic evaluation function according to \gleichung \ref{eq:AStarHeuristic}, 

\begin{equation}
f(v_j) = g(v_j) + h(v_j)
\label{eq:AStarHeuristic}
\end{equation}

where $g(v_j)$ is the cost of the path from the start node $v_1$ to the current node $v_j$ and $h(v_j)$ is an estimate of the remaining cost from the current node $v_j$ to the end node $v_N$. Thus f($v_j$) approximates the total cost of the path from $v_1$ via $v_j$ to $v_N$. For $g(v_j)$ the cost function from \gleichung \ref{eq:JAll} from $v_1$ to $v_j$ can be used directly. When selecting $h(v_j)$, note that $h(v_j)$ must not overestimate the actual cost from $v_j$ to $v_N$. Therefore, $h(v_j)$ uses the cost function from \gleichung \ref{eq:JAll} from $v_j$ to the target node $v_N$ using a vanishing detection probability ($P_\mathrm{D}(v_j)=\SI{0}{\percent}$ for all nodes from $j+1$ to $N$). Thus, the A* algorithm defines an iterative process, which begins at $v_1$ and finds $\vec{x}_\mathrm{oap}$ to the desired end node $v_N$.

\begin{figure}[b]
	\centering
	\tikzsetnextfilename{searchSpace_plot}\pgfplotsset{compat=newest}
\pgfplotsset{%
	xlabel style={at={(axis description cs:0.5,-0.13)}}, 
	ylabel style={at={(axis description cs:-0.10,0.5)}},
	y tick label style={/pgf/number format/fixed}
}
\definecolor{TUMBlue}{RGB}{0,101,189}
\definecolor{TUMOrange}{RGB}{227,114,34}
\definecolor{TUMGray2}{RGB}{127,127,127}
\begin{tikzpicture}
	\footnotesize
	\clip(-4,-1.9)rectangle(4.25,2.3);
	\node at (0,0) {
	\setlength\figureheight{3.5cm} 
	\setlength\figurewidth{7cm}
	\input{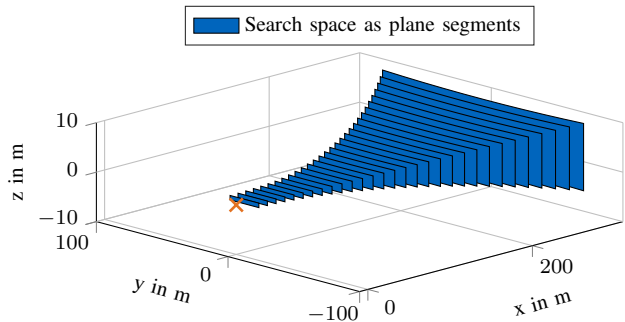}
	};
\fill[fill=white] (-2.5,-2.3) rectangle (-1,-1.85);
\fill[fill=white] (2.5,-2.15) rectangle (3.65,-1.8);
\node[align=left, anchor=center, rotate=15] at (3.15,-1.65) {\footnotesize x in m};
\node[align=left, anchor=center, rotate=-15] at (-1.9,-1.5) {\footnotesize y in m};
%
\end{tikzpicture}
	\caption{Valid search space for design class EKA 1 B according to the German Motorway Construction Guideline \cite{RAA} with minimum curve radius of $R_\mathrm{min} = \SI{720}{\meter}$, minimum crest radius of $H_\mathrm{K,min} = \SI{10 000}{\meter}$ and minimum hollow radius of $H_\mathrm{W,min} = \SI{5 700}{\meter}$. The position of the \ego is marked with an orange cross.}
	\label{fig:searchSpace}
\end{figure}

In order to increase the efficiency of the algorithm, the number of nodes in the 3D grid can be limited to the relevant nodes. Only approach paths to the \ego from the front are considered, therefore all nodes behind the \ego can be removed from the grid. In addition, a maximum distance in x-direction (vehicle longitudinal direction) of $\SI{300}{\meter}$ is defined, which corresponds approximately to the maximum range of sensors in the automotive industry. In addition, this paper considers the use case of German motorways. In Germany, the design of motorways is clearly regulated in the German Motorway Construction Guideline \cite{RAA}. This means, for example, that minimum curve radii as well as minimum radii of crest and hollow are specified. Based on the minimum values of these parameters, only nodes within a theoretically possible motorway course need to be considered. All other nodes can be removed from the grid. For the design class EKA 1 B standard cross section 43,5 (four lanes in each driving direction) according to the German Motorway Construction Guideline \cite{RAA}, the valid search space shown by section planes in \abbildung \ref{fig:searchSpace} is obtained.

In summary, the optimization algorithm calculates an optimal approaching path $\vec{x}_\mathrm{oap}$ based on a grid with detection probabilities and a cost function. This represents the optimal relative approach of a challenger to the \ego so that it can perceive the challenger as poorly as possible.  
%

\subsection{Scenario Transformation}
\label{sub:scenarioTrafo}
If $\vec{x}_\mathrm{oap}$ from the challenger to the \ego is calculated with the A* algorithm, then this represents the optimal course that the challenger must choose when the \ego is stationary. Since in reality both vehicles move, the trajectories of the \ego and the challenger have to be calculated from $\vec{x}_\mathrm{oap}$, which we call scenario transformation. If both vehicles execute the calculated trajectory (which also defines the street geometry of the scenario), $\vec{x}_\mathrm{oap}$ results as a relative movement between the vehicles. To achieve this, we need to make assumptions about the velocities of the \ego $v_\mathrm{ego}$ (drives with recommended speed on German motorways) as well as the challenger $v_\mathrm{ch}$ (slow moving vehicle) and set maximum allowable values for the pitch and yaw angles and rates of both vehicles, respectively (\tabelle\ref{tab:assumptions}).

\begin{figure}[b]
	\centering
	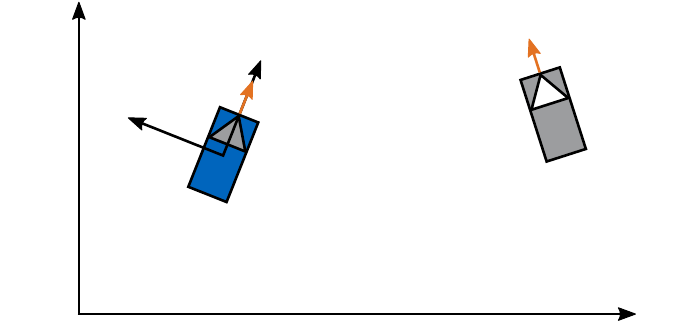
	\caption{Representation of the relative and global coordinate system in time step $i$.}
	\label{fig:CoSyA}
	\centering
	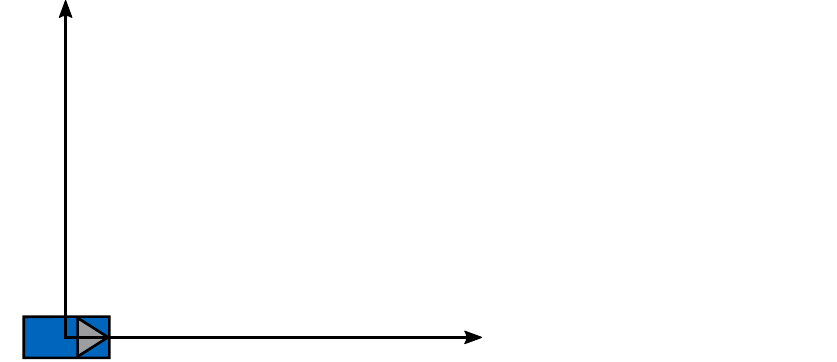
	\caption{Geometric relationships for calculating the trajectories in time step $i$, resulting in the optimal approximation path.}
	\label{fig:CoSyB}
\end{figure}

For a better transformation into a valid road profile, $\vec{x}_\mathrm{oap}$ is interpolated so that the data for the defined relative speed between \ego and challenger is available for each time step $i$ ($\Delta t = \SI{0.01}{\second}$). In addition, $\vec{x}_\mathrm{oap}$ is smoothed because no driving physics is taken into account during path optimization and the path can therefore contain corners. A global and a relative coordinate system are used for the calculation of the trajectories (\abbildung \ref{fig:CoSyA}). Subsequently, $\vec{x}_\mathrm{oap}$ must be converted from the relative to the global coordinate system. The aim of the algorithm is to determine the changes in position in every time step $i$ of the \ego $\Delta \vec{x}_{\mathrm{ego},i}$ and of the challenger $\Delta \vec{x}_{\mathrm{ch},i}$ so that $\vec{x}_\mathrm{ch,rel}(t)$ corresponds as closely as possible to $\vec{x}_\mathrm{oap}$ (\abbildung \ref{fig:CoSyB}).


\setlength{\tabcolsep}{2pt} 
\begin{table}[t]
	\centering
	\caption{Assumptions and maximum values used for the scenario transformation algorithm}
	\begin{tabular}{c r l l} 
		\toprule
		Symbol & Value & Unit & Description \\ 
		\midrule
		$v_\mathrm{ego}$		 	& \SI{130}{} 	& \SI{}{\kilo\metre\per\hour} 	& \ego velocity \\ 
		$v_\mathrm{ch}$  			& \SI{80}{}  	& \SI{}{\kilo\metre\per\hour} 	& challenger velocity \\
		$\dot{\Psi}_\mathrm{max}$ 	& \SI{0.22}{} 	& \SI{}{\radian\per\second} 	& max. yaw rate of both vehicles \\
		$\Delta\Psi_\mathrm{max}$ 	& \SI{0.21}{} 	& \SI{}{\radian}				& max. yaw angle difference between both \\
									& 				& 								& vehicles to increase stability of algorithm \\
		$\dot{\Theta}_\mathrm{max}$ & \SI{0.22}{} 	& \SI{}{\radian\per\second} 	& max. pitch rate of both vehicles \\
		$\Theta_\mathrm{max}$ 		& \SI{6.0}{} 	& \SI{}{\percent} 				& max. inclination of highways \cite{RAA} \\
		\bottomrule
	\end{tabular}
	\label{tab:assumptions}
\end{table}
\setlength{\tabcolsep}{6pt} 


At the beginning of the calculation, the \ego starts at the origin of the global coordinate system and the challenger at $v_1$ of $\vec{x}_\mathrm{oap}$ with their defined velocities in positive $x$-direction of the global coordinate system. It is important that the \ego is at some stage at the starting position of the challenger because the trajectories of both vehicles must be within the same road. Until this point is reached ($x_\mathrm{ego,glob} = x_\mathrm{ch,start,glob}$), the road geometry between the \ego and the challenger can be specified by the \ego movement. After this point, the \ego must follow the road course that the challenger has already defined. Thus, when the \ego reaches the starting position of the challenger, a change in the calculation method of the optimal road course arises. Therefore, the calculation is divided into these two areas (\algo \ref{alg:scenarioTransformationStart} and \ref{alg:scenarioTransformationEnd}). 

\DecMargin{2.5mm}
\begin{algorithm}[t]
	\SetKw{KwCalc}{calc}
	\SetKw{KwCheck}{check}
	\DontPrintSemicolon
	\KwIn{$\vec{x}_\mathrm{oap}$, $\Delta \vec{x}_{\mathrm{ego},i}$, $v_1$, $v_N$}
	\Parameter{$v_\mathrm{ego}$, $v_\mathrm{ch}$, $\Delta t$, $\dot{\Psi}_\mathrm{max}$, $\Delta\Psi_\mathrm{max}$, $\dot{\Theta}_\mathrm{max}$, $\Theta_\mathrm{max}$}
	\KwOut{$\vec{x}_\mathrm{ego}(t)$, $\vec{x}_\mathrm{ch}(t)$ and road geometry up to $v_1$}
	$t_i = 0$\;
	\While{$x_\mathrm{ego,glob} \leq x_\mathrm{ch,start,glob}$}{
		\KwCalc{$\Delta \vec{x}_\mathrm{oap,i}$			\tcp*{required relative path}}
		\KwCalc{$\Delta \vec{x}_{\mathrm{ch,req},i}$	\tcp*{required challenger path}}
		\KwCheck{$\dot{\Psi}_\mathrm{max}$, $\Delta\Psi_\mathrm{max}$, $\dot{\Theta}_\mathrm{max}$, $\Theta_\mathrm{max}$ \tcp*{of ch.}}
		\KwCalc{$\Delta \vec{x}_{\mathrm{ch},i}$		\tcp*{actual challenger pos. change}}
		$t_{i} \leftarrow t_i + \Delta t$		
}
\caption{Calculation of trajectories until \ego reaches the start pos. of the challenger at $v_1$.}
\label{alg:scenarioTransformationStart}
\end{algorithm}
\IncMargin{2.5mm}
\DecMargin{2.5mm}
\begin{algorithm}[!t]
	\SetKw{KwCalc}{calc}
	\SetKw{KwCheck}{check}
	\DontPrintSemicolon
	\KwIn{$\vec{x}_\mathrm{oap}$, $x_\mathrm{sp}$, $v_1$}
	\Parameter{$v_\mathrm{ego}$, $v_\mathrm{ch}$, $\Delta t$, $\dot{\Psi}_\mathrm{max}$, $\Delta\Psi_\mathrm{max}$, $\dot{\Theta}_\mathrm{max}$, $\Theta_\mathrm{max}$}
	\KwOut{$\vec{x}_\mathrm{ego}(t)$, $\vec{x}_\mathrm{ch}(t)$ and road geometry after $v_1$}
	$t_i = 0$\;
	\While{$x_\mathrm{ego,glob} \leq x_\mathrm{ch,glob}$}{
		\KwCalc{$\Delta \vec{x}_{\mathrm{ego,req},i}$			\tcp*{required ego pos. change}}
		\KwCheck{$\dot{\Psi}_\mathrm{max}$, $\Delta\Psi_\mathrm{max}$, $\dot{\Theta}_\mathrm{max}$, $\Theta_\mathrm{max}$ \tcp*{of ego}}
		\KwCalc{$\Delta \vec{x}_{\mathrm{ego},i}$				\tcp*{actual ego pos change}}		
		\KwCalc{$\vec{x}_{\mathrm{ch,rel},i}$			\tcp*{relative challenger position}}
														\tcp*{due to pos change of ego}
		\KwCalc{arguments of while-loop of \algo \ref{alg:scenarioTransformationStart}}		
	}
	\caption{Calculation of trajectories after \ego reaches the start pos. of the challenger at $v_1$.}
	\label{alg:scenarioTransformationEnd}
\end{algorithm}
\IncMargin{2.5mm}

In the first part of the calculation (\algo \ref{alg:scenarioTransformationStart}), the position change of the \ego $\Delta \vec{x}_{\mathrm{ego},i}$ in each time step $i$ is determined between the starting point of the \ego at $v_N$ and the starting point of the challenger at $v_1$ by means of an interpolation with a smoothing spline. The calculations in \algo \ref{alg:scenarioTransformationStart} consist of geometric correlations that are shown simplified in 2D in \abbildung \ref{fig:CoSyA} and \ref{fig:CoSyB}. For a detailed formulation of the equations, the interested reader is referred to \cite[chap. 4.5]{Lanz.2019}. At the end of the algorithm, the trajectories of both vehicles can be calculated from the constant speeds and the calculated paths $\Delta \vec{x}_{\mathrm{ego},i}$ and $\Delta \vec{x}_{\mathrm{ch},i}$. In the second part of the calculation (\algo \ref{alg:scenarioTransformationEnd}), the path of \ego $\Delta \vec{x}_{\mathrm{ego},i}$ must be within the road boundaries already defined by the challenger and is therefore given within limits. 

In summary, the scenario transformation uses $\vec{x}_\mathrm{oap}$, which represents the relative movement between the challenger and the \egowo, and calculates the necessary trajectories of the two vehicles under certain assumptions, so that $\vec{x}_\mathrm{oap}$ results as a relative movement. From the trajectories, the required road geometry can be derived as a surrounding envelope with the width of the standard cross section used. 

The calculated trajectory of the challenger as well as the derived road geometry represent the input for the most critical scenario with respect to the sensor setup of the \ego and can be executed in virtual simulation. The conversion into a simulation compatible format and the execution of the simulation are not part of this publication.

%
\section{RESULTS}
\label{sec:results}
This chapter first presents the results of the calculation of $\vec{x}_\mathrm{oap}$ using the A* algorithm (Section \ref{sub:optiAlgo}), and the second part describes the results of the scenario transformation (Section \ref{sub:scenarioTrafo}).

\subsection{Optimal Approaching Path with A*}
\label{sec:ASres}

The results shown here are calculated on the basis of the coarse grid (\tabelle \ref{tab:increments}), resulting in a total number of 8\,926 nodes. For the calculation of $P_\mathrm{D}$ of the grid according to \cite{Ponn.2019b}, good weather conditions and a passenger car as the challenger are used. The \ego is in the origin, so $\vec{v}_N = (\SI{0}{\metre}, \SI{0}{\metre}, \SI{0.5}{\metre})$. The challenger is outside the sensor coverage ($P_D = \SI{0}{\percent}$) at $\vec{v}_1 = (\SI{300}{\metre}, \SI{-20}{\metre}, \SI{2}{\metre})$, so the algorithm can select the optimal entry point in the sensor coverage. In the following, the influence of $k_\mathrm{J}$ of the cost function from \gleichung \ref{eq:JAll} will be examined in more detail. $\vec{x}_\mathrm{oap}$ for three different values of $k_\mathrm{J}$ are shown for the $x$-$y$-$\mathrm{plane}$ in \abbildung \ref{fig:approachingPathAudi_pathplot}. A detailed representation of the $x$-$z$-$\mathrm{plane}$ is not provided because $k_\mathrm{J}$ has no significant influence on it. Also, the choice of the grid size has only a small influence \cite[chap. 5]{Lanz.2019}.

\begin{table}[b]
	\centering
	\caption{Used grid sizes with $x_{\mathrm{min}} = \SI{0}{m}$, $x_{\mathrm{max}} = \SI{300}{m}$ and $y_{\mathrm{min}}$, $y_{\mathrm{max}}$, $z_{\mathrm{min}}$ and $z_{\mathrm{max}}$ according to \abbildung \ref{fig:searchSpace}.}
	\begin{tabular}{c c c c} 
		\toprule
		Grid & $\Delta x$ in \SI{}{\meter} & $\Delta y$ in \SI{}{\meter} & $\Delta z$ in \SI{}{\meter} \\ 
		\midrule
		fine	& \SI{1}{} 	& \SI{1}{} 	& \SI{0.2}{} \\ 
		middle 	& \SI{2}{} 	& \SI{2}{} 	& \SI{0.4}{} \\
		coarse 	& \SI{4}{} 	& \SI{4}{} 	& \SI{0.8}{} \\
		\bottomrule
	\end{tabular}
	\label{tab:increments}
\end{table}

\begin{figure}[b]
	\centering
	\tikzsetnextfilename{approachingPathAudi_pathplot}\pgfplotsset{compat=newest}
\pgfplotsset{%
	xlabel style={at={(axis description cs:0.5,-0.13)}}, 
	ylabel style={at={(axis description cs:-0.10,0.5)}},
	y tick label style={/pgf/number format/fixed}
}
\usetikzlibrary{positioning}
\tikzset{>=latex}
\definecolor{TUMBlue}{RGB}{0,101,189}
\definecolor{TUMOrange}{RGB}{227,114,34}
\definecolor{TUMGreen}{RGB}{162, 173, 0}
\definecolor{TUMIvory}{RGB}{218, 215, 203}
\definecolor{TUMDiag7}{RGB}{152, 198, 234}
\definecolor{TUMDiag14}{RGB}{249, 186, 0}

\begin{tikzpicture}
	\footnotesize
	\clip(-4,-2.2)rectangle(4.5,2.5);
	\node at (0,0) {
	\setlength\figureheight{3.5cm} 
	\setlength\figurewidth{7cm}
	\input{approachingPathAudi_pathplot_matlab.tikz} 
	};
%
%
\filldraw[fill=white, draw=black] (-2,2.15) rectangle (3.1,2.45);
\node[circle,draw=TUMBlue,line width=0.4mm, fill=white, inner sep=0pt,minimum size=5pt,label=east:{k$_J = 0.1$}] (a) at (-1.7,2.3) {};
\node[circle,draw=TUMGreen,line width=0.4mm, fill=white, inner sep=0pt,minimum size=5pt,label=east:{k$_J = 0.25$},right=1.4cm of a] (b) {};
\node[circle,draw=TUMOrange,line width=0.4mm, fill=white, inner sep=0pt,minimum size=5pt,label=east:{k$_J = 0.5$},right=1.5cm of b] (c) {};
\filldraw[fill=white, draw=black] (-0.75,1.55) rectangle (3.65,1.95);
\node[rectangle,draw=TUMIvory!65!,line width=0.4mm, fill=TUMIvory!65!, inner sep=0pt,minimum size=5pt,label=east:{Radar}] (d) at (-0.45,1.75) {};
\node[rectangle,draw=TUMDiag14!30!,line width=0.4mm, fill=TUMDiag14!30!, inner sep=0pt,minimum size=5pt,label=east:{Camera},right=1.2cm of d] (e) {};
\node[rectangle,draw=TUMDiag7!30!,line width=0.4mm, fill=TUMDiag7!30!, inner sep=0pt,minimum size=5pt,label=east:{Lidar},right=1.4cm of e] (f) {};

\node [fill=white,thick,minimum width=0.5cm,minimum height=0.3cm] (rect) at (1.5,-1.1) {\footnotesize Road boundaries};
\draw [->, thick] (2.5,-1) -- (2.9,-0.6);
\node [fill=white,thick,minimum width=0.2cm,minimum height=0.2cm] (rect) at (3.65,0.75) {\footnotesize $\vec{v}_1$};
\draw [->, thick] (3.65,0.5) -- (3.75,-0.05);
\node [fill=none,thick,minimum width=0.1cm,minimum height=0.1cm] (rect) at (-2.4,-0.65) {\footnotesize $\vec{v}_N$};
\draw [->, thick] (-2.5,-0.5) -- (-2.57
,0.35);
\end{tikzpicture}
	\caption{Optimal approaching path $\vec{x}_\mathrm{oap}$ with different weighting factors $k_\mathrm{J}$.}
	\label{fig:approachingPathAudi_pathplot}
\end{figure}

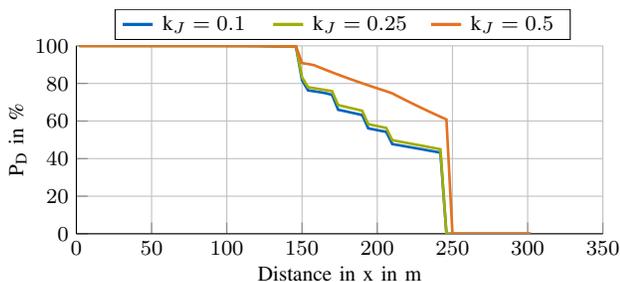
\begin{figure}[t]
	\centering
	\tikzsetnextfilename{approachingPathAudi_detectionplot}\pgfplotsset{compat=newest}
\pgfplotsset{%
	xlabel style={at={(axis description cs:0.5,-0.13)}}, 
	ylabel style={at={(axis description cs:-0.08,0.5)}},
	y tick label style={/pgf/number format/fixed}
}
\definecolor{TUMBlue}{RGB}{0,101,189}
\definecolor{TUMOrange}{RGB}{227,114,34}
\definecolor{TUMGray2}{RGB}{127,127,127}
\definecolor{TUMGreen}{RGB}{162, 173, 0}
\begin{tikzpicture}
	\footnotesize
	\clip(-4,-1.65)rectangle(4.2,2);
	\node at (0,0) {
	\setlength\figureheight{2.5cm} 
	\setlength\figurewidth{7cm}
%
\definecolor{mycolor1}{rgb}{0.00000,0.39608,0.74118}%
\definecolor{mycolor2}{rgb}{0.63529,0.67843,0.00000}%
\definecolor{mycolor3}{rgb}{0.89020,0.44706,0.13333}%
\begin{tikzpicture}

\begin{axis}[%
width=\figurewidth,
height=\figureheight,
at={(0\figurewidth,0\figureheight)},
scale only axis,
xmin=0,
xmax=350,
xlabel={Distance in x in m},
xmajorgrids,
ymin=0,
ymax=100,
ylabel={$\text{P}_\text{D}\text{ in \%}$},
ymajorgrids,
axis x line*=bottom,
axis y line*=left,
legend style={legend cell align=left,align=left,draw=white!15!black}
]
\addplot [color=mycolor1,solid,line width=1.0pt,forget plot]
  table[row sep=crcr]{%
2	100\\
6	100\\
10	100\\
14	100\\
18	100\\
22	100\\
26	100\\
30	100\\
34	100\\
38	100\\
42	100\\
46	100\\
50	100\\
54	100\\
58	100\\
62	100\\
66	100\\
70	100\\
74	99.9998822937318\\
78	99.9993989870914\\
82	99.9986105781094\\
86	99.9974580525454\\
90	99.9956585015241\\
94	99.989228160159\\
98	99.9808548377769\\
102	99.9756276099513\\
106	99.9697049210071\\
110	99.9623544428976\\
114	99.9502620148215\\
118	99.9352792055083\\
122	99.9179431389477\\
126	99.8363200659367\\
130	99.7999158486891\\
134	99.7372670952285\\
138	99.66576046926\\
142	99.5837426442637\\
146	99.4919000287514\\
150	81.7066026595183\\
154	76.3068781449791\\
158	75.8314701195376\\
162	75.3452607180236\\
166	74.759577626369\\
170	73.9332997266308\\
174	65.9922042062276\\
178	65.3238669361764\\
182	64.5965551819123\\
186	63.909126692554\\
190	63.1775206375798\\
194	56.1005504264214\\
198	55.4811090282556\\
202	54.8336150904889\\
206	54.239554858722\\
210	47.7214622550443\\
214	47.1859572784535\\
218	46.6184211145567\\
222	46.0537390931647\\
226	45.4813441765823\\
230	44.9423777576574\\
234	44.320839576549\\
238	43.7710668369092\\
242	43.177987854211\\
246	0\\
250	0\\
254	0\\
258	0\\
262	0\\
266	0\\
270	0\\
274	0\\
278	0\\
282	0\\
286	0\\
290	0\\
294	0\\
298	0\\
302	0\\
302	0\\
};
\addplot [color=mycolor2,solid,line width=1.0pt,forget plot]
  table[row sep=crcr]{%
2	100\\
6	100\\
10	100\\
14	100\\
18	100\\
22	100\\
26	100\\
30	100\\
34	100\\
38	100\\
42	100\\
46	100\\
50	100\\
54	100\\
58	100\\
62	100\\
66	100\\
70	100\\
74	99.9999211859094\\
78	99.9995905762053\\
82	99.9989840584789\\
86	99.997986358172\\
90	99.9964687609811\\
94	99.9908346654367\\
98	99.9836296840794\\
102	99.9786660107371\\
106	99.9726823771564\\
110	99.9656896997191\\
114	99.9544317690187\\
118	99.940384648281\\
122	99.9240264695579\\
126	99.8601773586854\\
130	99.8258092351061\\
134	99.768839397275\\
138	99.7036115691318\\
142	99.6279385013334\\
146	99.5427410945542\\
150	83.2494095134575\\
154	77.9859117040224\\
158	77.4712940366546\\
162	76.9421663680902\\
166	76.4222749341279\\
170	75.8856165435682\\
174	68.4708614614445\\
178	67.7456630912095\\
182	66.9798016773464\\
186	66.2282082920188\\
190	65.4536410170552\\
194	58.3252789818123\\
198	57.661499624993\\
202	56.9767880058132\\
206	56.3092871985908\\
210	49.7713544746592\\
214	49.2053033037875\\
218	48.584204011976\\
222	47.9848157605766\\
226	47.3883382566653\\
230	46.8185842190807\\
234	46.1803929712577\\
238	45.5853620358637\\
242	44.9628522387034\\
246	0\\
250	0\\
254	0\\
258	0\\
262	0\\
266	0\\
270	0\\
274	0\\
278	0\\
282	0\\
286	0\\
290	0\\
294	0\\
298	0\\
302	0\\
302	0\\
};
\addplot [color=mycolor3,solid,line width=1.0pt,forget plot]
  table[row sep=crcr]{%
2	100\\
6	100\\
10	100\\
14	100\\
18	100\\
22	100\\
26	100\\
30	100\\
34	100\\
38	100\\
42	100\\
46	100\\
50	100\\
54	100\\
58	100\\
62	100\\
66	100\\
70	100\\
74	100\\
78	100\\
82	100\\
86	100\\
90	99.9996656188291\\
94	99.9985446937914\\
98	99.9964267736545\\
102	99.9935932284788\\
106	99.9897909943666\\
110	99.9849127101254\\
114	99.9779869394361\\
118	99.9677840677317\\
122	99.9550168404016\\
126	99.9394767067875\\
130	99.9209772719749\\
134	99.8946412500714\\
138	99.8588408794121\\
142	99.816839530903\\
146	99.7683764910963\\
150	90.8050228401444\\
154	90.3486675815704\\
158	89.7115078195272\\
162	88.4098541009514\\
166	87.1394679096743\\
170	85.8988575601544\\
174	84.6866607280474\\
178	83.5015859289172\\
182	82.3424437523536\\
186	81.2081098742705\\
190	80.0956979173636\\
194	79.0002961547709\\
198	77.9270560660148\\
202	76.8750910402239\\
206	75.8435666310404\\
210	74.7194942382411\\
214	73.0645649939021\\
218	71.4399890488446\\
222	69.8446624987692\\
226	68.2775405873654\\
230	66.7376335547012\\
234	65.2240028435965\\
238	63.7357576275687\\
242	62.2720516281807\\
246	60.8320801933219\\
250	0\\
254	0\\
258	0\\
262	0\\
266	0\\
270	0\\
274	0\\
278	0\\
282	0\\
286	0\\
290	0\\
294	0\\
298	0\\
302	0\\
};
\end{axis}
\end{tikzpicture}%
	};
%
%
\filldraw[fill=white, draw=black] (-2.55,1.6) rectangle (3.5,2);
\draw[TUMBlue, very thick] (-2.45,1.8) --  (-2.05,1.8) node[label={[label distance=-0.1cm, black]east:{k$_J = 0.1$}}] (d){};
\draw[TUMGreen, very thick] (-0.45,1.8) --  (-0.05,1.8) node[label={[label distance=-0.1cm, black]east:{k$_J = 0.25$}}] (e){};
\draw[TUMOrange, very thick] (1.65,1.8) --  (2.05,1.8) node[label={[label distance=-0.1cm, black]east:{k$_J = 0.5$}}] (f){};
\end{tikzpicture}
	\caption{Detection probability $P_\mathrm{D}$ with different weighting factors $k_\mathrm{J}$. The lengths of the path are summarized in \tabelle \ref{tab:oapLength}. The high detection probabilities $P_\mathrm{D}$ of almost \SI{100}{\percent} are due to multiple overlapping sensors and the assumed good weather conditions. This can change in adverse weather conditions \cite{Ponn.2019b}.}
	\label{fig:approachingPathAudi_detectionplot}
\end{figure}

With varying $k_\mathrm{J}$, $\vec{x}_\mathrm{oap}$ changes significantly. For $k_\mathrm{J} = 0.5$ (high distance costs) the algorithm chooses the shortest possible path to the \ego and $P_\mathrm{D}$ of the nodes has only a negligible effect. With lower $k_\mathrm{J}$, more and more focus is placed on a lower $P_\mathrm{D}$ and a longer path is accepted. \abbildung \ref{fig:approachingPathAudi_detectionplot} shows $P_\mathrm{D}$ as a function of the $x$-distance. It can be seen that smaller $k_\mathrm{J}$ lead to lower $P_\mathrm{D}$. However, the total length of the approaching path is inversely proportional (\tabelle \ref{tab:oapLength}). 

\begin{table}[t]
	\centering
	\caption{Length of the optimal approaching path $L(\vec{x}_\mathrm{oap})$ with different weighting factors $k_\mathrm{J}$.}
	\begin{tabular}{c c c c} 
		\toprule
		$k_\mathrm{J}$ 				& \SI{0.1}{} 			& \SI{0.25}{} 			& \SI{0.5}{} \\ 
		\midrule
		$L(\vec{x}_\mathrm{oap})$	& \SI{343.9}{\meter} 	& \SI{325.0}{\meter} 	& \SI{308.4}{\meter} \\ 
		\bottomrule
	\end{tabular}
	\label{tab:oapLength}
\end{table}

As a consequence of these results, it can be concluded that small values of $k_\mathrm{J}$ are preferred for the creation of worst case scenarios regarding the sensor coverage of automated vehicles. 

\subsection{Scenario Transformation Results}
\label{subsec:SceTrans}

\begin{figure}[t]
	\centering
	\tikzsetnextfilename{generateRoadLayout_plot}\pgfplotsset{compat=newest}
\pgfplotsset{%
	xlabel style={at={(axis description cs:0.5,-0.13)}}, 
	ylabel style={at={(axis description cs:-0.10,0.5)}},
	y tick label style={/pgf/number format/fixed}
}
\definecolor{TUMBlue}{RGB}{0,101,189}
\definecolor{TUMOrange}{RGB}{227,114,34}
\definecolor{TUMGray2}{RGB}{127,127,127}
\tikzset{>=latex}
\begin{tikzpicture}
	\footnotesize
	\clip(-3.5,-1.6)rectangle(3.9,1.7);
	\node at (0,0) {
	\setlength\figureheight{2.5cm} 
	\setlength\figurewidth{7cm}
	\input{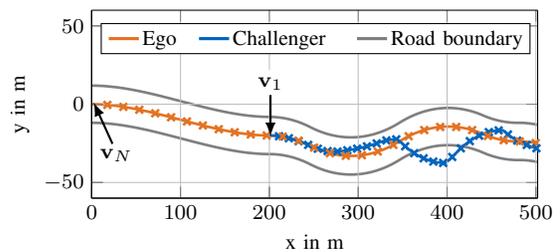}
	};
%
%
\filldraw[fill=white, draw=black] (-2.3,1.05) rectangle (3.4,1.45);
\draw[TUMOrange, very thick] (-2.2,1.25) -- (-1.8,1.25);
\node[align=left, anchor=west] at (-1.85,1.23) {\footnotesize Ego};
\draw[TUMBlue, very thick] (-1,1.25) -- (-0.6,1.25);
\node[align=left, anchor=west] at (-0.65,1.23) {\footnotesize Challenger};
\draw[TUMGray2, very thick] (1.1,1.25) -- (1.5,1.25);
\node[align=left, anchor=west] at (1.45,1.23) {\footnotesize Road boundary};
\node [fill=white,thick,minimum width=0.2cm,minimum height=0.2cm] (rect) at (-0.05,0.65) {\footnotesize $\vec{v}_1$};
\draw [->, thick] (-0.05,0.5) -- (-0.05,-0.05);
\node [fill=none,thick,minimum width=0.1cm,minimum height=0.1cm] (rect) at (-2.1,-0.3) {\footnotesize $\vec{v}_N$};
\draw [->, thick] (-2.15,-0.15) -- (-2.4,0.4);
\end{tikzpicture}
	\caption{Calculated optimal road geometry to generate the $\vec{x}_\mathrm{oap}$ between the two vehicles. From $\vec{v}_1$, i.e. from \SI{200}{\metre}, the road can theoretically also be defined around the challenger.}
	\label{fig:roadLayout}
\end{figure}

To demonstrate the results of the scenario transformation, we use the fine grid from \tabelle \ref{tab:increments}, a weighting factor of $k_\mathrm{J} = 0.25$, and a starting point of the challenger at $\vec{v}_1 = (\SI{200}{\metre}, \SI{-20}{\metre}, \SI{2}{\metre})$. This results in a scenario duration of \SI{14.06}{\second} with the assumptions made in \tabelle \ref{tab:assumptions}. The \ego covers a distance of \SI{507.7}{\meter} and the challenger \SI{312.4}{\meter}. Again, the visualization is reduced to the $x$-$y$-$\mathrm{plane}$ (\abbildung \ref{fig:roadLayout}). This shows the course of the road when both vehicles move, so that $\vec{x}_\mathrm{oap}$ is obtained as the relative movement between the two vehicles. 

\abbildung \ref{fig:roadLayout} shows the road boundaries as twice the width of the possible offset in $y$-direction between two vehicles for a standard cross-section of 43,5. The reference represents the \ego that is always in the middle of the boundaries. From the point of view of the \egowo, the challenger can be to the left and to the right with the highest possible offset to still be on the shared street layout. The highest possible offset in the $y$-direction for the standard cross section 43,5 is \SI{10.875}{\meter}. 

While in the $x$-$z$-$\mathrm{plane}$ no exceedances of the limits occur \cite[p. 78]{Lanz.2019}, in \abbildung \ref{fig:roadLayout} for the $x$-$y$-$\mathrm{plane}$ it can be seen that the challenger is outside the road limits in the region of $x=\SI{400}{\meter}$. This means that in this area no road can be defined according to \cite{RAA} in order to maintain $\vec{x}_\mathrm{oap}$ between the two vehicles. Apart from these short deviations, however, the procedure developed is suitable for defining the optimal road geometry for particularly challenging scenarios for the perception module of an automated vehicle. Approaches to overcome the limitations of the scenario transformation algorithm are discussed in Section \ref{sec:discussion}.



\section{DISCUSSION}
\label{sec:discussion}


For a more efficient calculation of $\vec{x}_\mathrm{oap}$, the grid can be reduced to 2D because the influence and effects in $z$-direction are negligible. The reason for this lies in the sensor models used to calculate $P_\mathrm{D}$ of the grid \cite{Ponn.2019b}. These show almost no dependence on the $z$-component. If, however, sensor models are used that consider a dependency in the $z$-direction, the 3D grid must be used instead. 

The most potential for improvement exists in scenario transformation (Section \ref{sub:scenarioTrafo}). \abbildung \ref{fig:roadLayout} shows that the boundary conditions cannot be met over the entire duration of the scenario. The following aspects describe approaches with which compliance with the boundary conditions can be improved:
\begin{itemize}
	\item Consideration of vehicle dynamics already during path search in the A* algorithm.
	\item No exact position specification of the \ego to the starting point $\vec{v}_1$ of the challengers (deviations within one road width tolerable).
	\item Currently the trajectories are calculated forward in time. This means that starting from the initial distance between the vehicles, both drive forward and get closer with $\vec{x}_\mathrm{oap}$. A reversal of the process into a reverse simulation can bring advantages, because the more complex calculation of the road geometry (both are on the same road) is already carried out at the beginning of the calculation. 
\end{itemize}

In addition, an adaptive velocity specification of the challenger may be necessary for the simulation execution of the generated critical scenarios. This is necessary if the \ego deviates from the assumed constant speed and the assumed relative speed between both vehicles must be re-established by the challenger.

\section{CONCLUSION}
\label{sec:conclusion}

This contribution addresses a novel method for the definition of system-specific challenging scenarios for the safety assessment of automated vehicles with the Operational Design Domain of highways. Based on a 3D grid of the sensor coverage with corresponding detection probabilities at the nodes of the grid, the A* algorithm calculates a worst-case approaching path between the \ego and the challenger, so that the automated vehicle perceives the challenger as poorly as possible. The static approaching path is translated into a dynamic scenario using scenario transformation (Section \ref{sub:scenarioTrafo}), which defines the challenger trajectory and road geometry. These scenarios represent the most critical scenarios from the point of view of the automated vehicle perception module and offer significant additional value in the safety assessment of automated driving on highways. 
In future work, visual obstructions caused by other objects can be considered. In addition, the generated scenario data will be exported directly to an OpenSCENARIO and OpenDRIVE file for automated connection to a simulation tool.

\section*{ACKNOWLEDGMENT AND CONTRIBUTIONS}

Thomas Ponn (corresponding author) initiated and wrote this paper. He was involved in all stages of development and primarily developed the research question as well as the concept. Thomas Lanz wrote his master thesis on sensor modeling and implemented the developed models during his thesis. Frank Diermeyer contributed to the conception of the research project and revised the paper critically for important intellectual content. He gave final approval of the version to be published and agrees to all aspects of the work. As a guarantor, he accepts responsibility for the overall integrity of the paper.

\bibliographystyle{./bibliography/IEEEtran} 
\bibliography{./bibliography/IEEEabrv,./bibliography/mybibfile}

\end{document}